\documentclass[11pt]{article}

\usepackage[margin=1in]{geometry}
\usepackage[utf8]{inputenc}
\usepackage[T1]{fontenc}
\usepackage{lmodern}
\usepackage{microtype}
\usepackage{amsmath, amssymb, amsthm}
\usepackage{graphicx}
\usepackage{booktabs}
\usepackage{multirow}
\usepackage{array}
\usepackage{xcolor}
\usepackage{xspace}
\usepackage{hyperref}
\usepackage{cleveref}
\usepackage{algorithm}
\usepackage{algpseudocode}
\usepackage{subcaption}
\usepackage{enumitem}
\usepackage{authblk}

\hypersetup{
    colorlinks=true,
    linkcolor=blue!60!black,
    citecolor=blue!60!black,
    urlcolor=blue!60!black
}

\newcommand{\nvfp}{\textsc{NVFP4}\xspace}

\newcommand{\nemotron}{Nemotron-3-Nano-30B-A3B\xspace}
\newcommand{\qwen}{Qwen3-4B-Thinking-2507\xspace}
\newcommand{\cka}{\mathrm{CKA}}
\newcommand{\hsic}{\mathrm{HSIC}}

\newcommand{\ckaqad}{CKA-QAD\xspace}

\title{\textbf{Beyond Output Matching:\\
Preserving Internal Geometry in \nvfp{} LLM Distillation}}

\author{Fangbo Tu}
\author{Junhua Zhao}
\author{Chi Liu}
\author{Xin Chen}
\author{Haifeng Wu}
\author{Jian Wan}
\author{Srinivasan Manoharan}
\affil{PayPal, Inc.}

\date{}

\begin{document}
\maketitle

\begin{abstract}
Demand for low-precision inference, including NVFP4-based approaches, has grown as large language models are increasingly deployed in latency and cost constrained production environments. Quantization-aware distillation (QAD) helps recover accuracy lost under low bit quantization by training a quantized student to match the output distribution of a frozen higher precision teacher via a KL-divergence loss. In this work, we first provide a representation level diagnosis of QAD: output matching alone can mask internal degradation, because many intermediate activation geometries can yield similar teacher-aligned logits. Using CKA, we show that KL-only QAD can reduce layerwise representational similarity relative to the BF16 teacher, with especially severe drift in RL-post-trained models. This drift correlates with downstream bottlenecks on reasoning and coding tasks, suggesting that low bit recovery requires preserving internal geometry rather than matching outputs alone. Motivated by this finding, we propose \textbf{\ckaqad{}}, a CKA-guided representational alignment method for \nvfp{} QAD and low bit LLM accuracy recovery. The method adds a lightweight regularizer that preserves internal representational geometry during distillation by aligning layerwise Gram matrices through CKA. Across Nemotron 3 Nano and \qwen{}, \ckaqad{} substantially improves representational alignment and improves downstream reasoning and coding accuracy with modest training overhead. Our findings position CKA-guided representational alignment as a practical complement to output matching for quantized LLM recovery.
\end{abstract}

\section{Introduction}
\label{sec:intro}

The rapid deployment of large language models (LLMs) in latency and memory constrained environments has accelerated the adoption of low precision inference formats. Among recent advances, \nvfp{}~\cite{xin2026qad}, a 4-bit floating point format with fine-grained block scaling and extended dynamic range, offers substantial throughput and memory benefits over FP8 and INT4. However, post training quantization (PTQ) to \nvfp{} often incurs non-negligible accuracy degradation, particularly for compact or reasoning specialized architectures. Quantization-aware distillation (QAD) has emerged as a highly effective remedy, aligning the quantized student's output distribution with its full precision teacher via KL divergence. Unlike quantization-aware training (QAT), QAD avoids replicating complex multi stage post training pipelines (e.g., supervised finetuning and reinforcement learning) and demonstrates remarkable stability across diverse model families~\cite{xin2026qad}.

Despite its success in recovering task-level accuracy, the prevailing QAD paradigm primarily focuses on matching output level probability distributions. This raises a critical question: \textit{does output alignment guarantee faithful internal representation recovery?} In this work, we systematically analyze the internal geometry of \nvfp{} quantized models during QAD and uncover a significant disconnect. We find that PTQ alone induces noticeable degradation in layerwise representational similarity, as quantified by Centered Kernel Alignment (CKA)~\cite{kornblith2019similarity}. Surprisingly, standard QAD optimization using only a KL divergence loss often \textit{exacerbates} this representational drift, particularly in models that undergo reinforcement learning (RL) post training. While KL-based distillation aligns the student's output distribution with the teacher's, the student's intermediate representations may still undergo unintended reorganization. This suggests that output level alignment can mask internal representational degradation that is not captured by the KL objective alone.

We hypothesize that preserving the teacher’s internal representational geometry is important for robust low bit generalization, especially on complex reasoning and coding tasks that rely on deep semantic hierarchies. To address this, we frame our contribution as CKA-guided representational alignment for NVFP4 QAD and low bit LLM accuracy recovery, and instantiate it as CKA-QAD, a lightweight extension to the QAD objective that explicitly regularizes intermediate activations using CKA similarity. By jointly optimizing output distribution alignment via KL divergence and internal representation alignment via CKA, our method discourages superficial mappings that satisfy the output level loss at the expense of representational fidelity. Crucially, CKA’s invariance to orthogonal transformations and isotropic scaling makes it less sensitive than pointwise feature losses to benign rotations and global scale changes in activation space, which is desirable for low bit activations whose distributions may shift during quantized training.

We evaluate \ckaqad{} on Nemotron 3 Nano, a hybrid Mamba--Transformer MoE, and \qwen{}, a compact Transformer reasoning model. Our key findings are threefold:
\begin{itemize}[leftmargin=*, nosep]
    \item \textbf{Representation Drift in QAD:} We quantify and visualize how KL-only distillation leads to reduced CKA similarity between the quantized student and the high precision teacher across both Transformer and Mamba blocks, with heavily RL-post-trained models exhibiting the most pronounced drift despite close output distribution alignment with the BF16 teacher.
    \item \textbf{CKA-Guided \nvfp{} QAD:} We introduce a layerwise CKA alignment loss for \nvfp{} QAD, restoring internal alignment to near BF16 levels while adding only 0.5\% step time overhead and 7.0\% peak VRAM overhead in the matched \qwen{} training profile.
    \item \textbf{Downstream Accuracy:} On \qwen{}, \ckaqad{} improves average accuracy over QAD (KL-only) on AIME25 (68.5\% to 72.3\%), GPQA-D (59.5\% to 61.1\%), and LiveCodeBench-v5 (57.9\% to 59.8\%).
\end{itemize}
These results suggest that efficient \nvfp{} recovery benefits from joint optimization of \textit{functional} and \textit{representational} alignment, motivating CKA-guided recovery as a practical direction for low bit LLM distillation.

The remainder of this paper is organized as follows. \Cref{sec:bg-related} reviews background and related work on \nvfp{} quantization, QAD, representation similarity, and distillation. \Cref{sec:method} introduces the \ckaqad{} objective and implementation. \Cref{sec:experiments} presents the experimental setup, main results, efficiency analysis, and model generalization study. \Cref{sec:disc,sec:concl} discuss limitations and conclude.

\section{Background and Related Work}
\label{sec:bg-related}

\subsection{NVFP4 quantization and quantization-aware distillation}
Low precision inference has become essential for deploying large language models (LLMs) under strict latency and memory budgets. The \nvfp{}~format~\cite{xin2026qad,alvarez2025nvfp4} extends NVIDIA’s microscaling FP4 design by using a smaller quantization block size (16 values, compared with 32 in MXFP4), per block FP8 E4M3 scale factors for fine-grained adaptation, and a second level per tensor FP32 scale to preserve dynamic range. This design reduces model memory footprint by approximately 1.8$\times$ relative to FP8 and can deliver 2--3$\times$ token-generation throughput speedups in practical LLM inference settings, making it attractive for production deployment.

Post training quantization (PTQ) applies \nvfp{} calibration without additional training, but often incurs non-negligible accuracy degradation on compact or reinforcement learning fine-tuned models~\cite{xin2026qad}. Quantization-aware distillation (QAD) addresses this by training a quantized student to match the output distribution of a full precision BF16 teacher. Let $z_T(x), z_S(x) \in \mathbb{R}^{|V|}$ denote the teacher and student logits for input $x$, where $V$ is the vocabulary. The standard temperature scaled distillation objective is
\begin{equation}
\mathcal{L}_{\mathrm{KD}} =
\mathbb{E}_{x \sim \mathcal{D}}
\left[
\tau^2 D_{\mathrm{KL}}\!\left(
\sigma\!\left(\frac{z_T(x)}{\tau}\right)
\,\Big\|\,
\sigma\!\left(\frac{z_S(x)}{\tau}\right)
\right)
\right],
\label{eq:kd}
\end{equation}
where $\sigma(\cdot)$ denotes the softmax function and $\tau$ is the distillation temperature. It is optionally combined with a small cross entropy term against the gold label. Compared with quantization-aware training (QAT)~\cite{nagel2021white}, QAD has practical advantages in modern LLM pipelines: it avoids replaying complex multi stage post training recipes such as supervised fine-tuning, RL, or model merging; it is comparatively robust to partial or synthetic calibration data; and it directly aligns the student with the BF16 teacher's output distribution. Despite these benefits, QAD's impact on the model's internal representational geometry remains underexamined.

\subsection{Representational similarity and Centered Kernel Alignment}
Understanding how internal representations evolve under compression or fine-tuning requires metrics that capture geometric similarity rather than pointwise activation equality. Centered Kernel Alignment (CKA)~\cite{kornblith2019similarity} is a standard tool for this purpose. Given matched activations $X \in \mathbb{R}^{N \times d_T}$ from the teacher and $Y \in \mathbb{R}^{N \times d_S}$ from the student over the same $N$ tokens, linear CKA is
\begin{equation}
\cka(X, Y) = \frac{\hsic(X, Y)}{\sqrt{\hsic(X,X)\,\hsic(Y,Y)}},
\quad
\hsic(X,Y) = \frac{1}{(N-1)^2}\,\mathrm{tr}(K_X H K_Y H),
\label{eq:cka}
\end{equation}
where $K_X = XX^\top$, $K_Y = YY^\top$, and $H = I - \tfrac{1}{N}\mathbf{1}\mathbf{1}^\top$ is the centering matrix. Equivalently, for centered activations, $\cka(X,Y)=\|Y^\top X\|_F^2 / (\|X^\top X\|_F\,\|Y^\top Y\|_F)$. CKA does not require $d_T=d_S$ and is invariant to orthogonal transformations and isotropic scaling, making it robust to feature rotations and magnitude shifts that commonly arise in quantized or fine-tuned networks. Although CKA has been widely used to compare representations across architectures, widths, depths, and random seeds, its role as a training-time guide for \nvfp{} QAD and low bit LLM accuracy recovery remains underexplored.

\subsection{Feature distillation versus logit distillation}
Knowledge distillation traditionally operates at the output level, aligning teacher and student probability distributions through KL divergence or temperature scaled cross entropy~\cite{hinton2015distilling}. To transfer richer structural knowledge, feature distillation methods align intermediate signals: FitNets~\cite{romero2014fitnets} minimizes feature regression losses between projected hidden states, Attention Transfer~\cite{zagoruyko2017attention} matches attention maps, and Similarity Preserving distillation~\cite{tung2019similarity} aligns instance wise affinity structure.

CKA-style representation matching has also been explored as a distillation signal. Prior work adapts CKA to transfer feature structures in BERT distillation~\cite{jung2023feature} and reexamines CKA as a representation alignment objective for visual knowledge distillation~\cite{zhou2024rethinking}. Our contribution is therefore not to introduce CKA as a generic distillation loss. Instead, we study CKA as a representation-preserving regularizer for \nvfp{} QAD, where the teacher and student share architecture but low bit quantization and KL-only recovery can still induce substantial internal geometry drift.

While effective in full precision regimes, these approaches face challenges in low bit recovery. First, \nvfp{}'s block wise E4M3 scaling and dynamic range compression induce non-stationary activation distributions that make direct activation matching brittle. Second, feature matching objectives that do not account for representational invariances may force the student toward activation equality rather than geometry preservation. Consequently, feature level distillation has seen limited adoption in 4-bit inference recovery, where logit level KL remains the de facto standard~\cite{xin2026qad}.

\subsection{The gap: output alignment versus internal geometry}
Current QAD pipelines operate under the implicit assumption that minimizing output level KL divergence also preserves internal representational fidelity. Our empirical analysis reveals this assumption to be flawed: optimizing KL alone can exacerbate representational drift, as quantified by lower CKA scores across Transformer and Mamba layers. This drift is particularly pronounced in RL-post-trained models, where the student may learn shortcut mappings that satisfy the distillation objective while reorganizing internal feature spaces in ways that harm complex reasoning and code generation.

The superposition hypothesis~\cite{elhage2022superposition} gives one lens on why this matters: neural networks can represent more features than dimensions by packing features into low-interference directions, making the geometry of activations functionally load bearing. The specific disconnect between output alignment and internal geometry in \nvfp{} QAD remains underexplored, as does the use of CKA as a practical alignment signal for 4-bit LLM recovery. Our novelty is to integrate CKA directly into the QAD objective as CKA-guided representational alignment for \nvfp{} QAD and low bit LLM accuracy recovery, showing that low bit distillation benefits from jointly optimizing functional alignment through KL and representational alignment through CKA.

\section{Method}
\label{sec:method}

\subsection{Overall Framework}
\label{sec:framework}

We build upon the standard Quantization-Aware Distillation (QAD) pipeline~\cite{xin2026qad}, which aligns a quantized \nvfp{} student model with a frozen BF16 teacher via KL divergence. While QAD effectively recovers output level functional accuracy, our analysis reveals that it often induces internal representational drift. To mitigate this, we augment the QAD objective with a Centered Kernel Alignment (CKA) regularization term that explicitly preserves the geometric structure of intermediate activations.

CKA-guided alignment encourages the quantized student to preserve the teacher’s internal representational geometry, rather than merely matching output logits. This is particularly critical in low-bit regimes, where quantization-induced scale shifts and clipping render pointwise feature matching (e.g., MSE-based alignment) brittle. In contrast, CKA is theoretically invariant to orthogonal transformations and feature scaling, making it robust to quantization artifacts. Consequently, CKA serves as a natural complement to KL-based QAD: while KL aligns the functional output distribution, CKA preserves the intermediate representational geometry essential for robust reasoning under aggressive quantization.

\subsection{CKA-Regularized Distillation Objective}
\label{sec:loss}

Let $z_t(x)$ and $z_s(x)$ denote the teacher and student logits for input $x$. To reduce the communication and memory cost of vocabulary level distillation, our implementation follows a teacher-top-$k$ variant of QAD. Let $\Omega_k(x)$ be the indices of the $k$ largest teacher logits, with $k=8192$ in our experiments. We define renormalized top-$k$ distributions
\[
    p_t^{(k)}(i|x) =
    \frac{\exp(z_{t,i}(x)/\tau)}
    {\sum_{j \in \Omega_k(x)} \exp(z_{t,j}(x)/\tau)},\quad
    p_s^{(k)}(i|x) =
    \frac{\exp(z_{s,i}(x)/\tau)}
    {\sum_{j \in \Omega_k(x)} \exp(z_{s,j}(x)/\tau)}
\]
for $i \in \Omega_k(x)$. The logit level distillation loss is:
\begin{equation}
    \mathcal{L}_{\mathrm{TopK}\text{-}\mathrm{KL}} = \mathbb{E}_{x \sim \mathcal{D}} \left[
    D_{\mathrm{KL}}\big(p_t^{(k)}(\cdot|x) \,\|\, p_s^{(k)}(\cdot|x)\big)
    \right].
\label{eq:method-kl}
\end{equation}
We introduce a layerwise CKA regularizer to align internal representations. Let $\mathcal{S}$ denote the set of selected decoder blocks, and let $H_t^{(\ell)}, H_s^{(\ell)} \in \mathbb{R}^{T \times B \times d}$ be the teacher and student block output tensors at layer $\ell$, where $T$ is sequence length, $B$ is batch size, and $d$ is hidden dimension. We flatten sequence and batch into $N=T B$ tokens, center features along the token dimension, and compute
\begin{equation}
    \rho_\ell =
    \frac{\|(\bar{H}_s^{(\ell)})^\top \bar{H}_t^{(\ell)}\|_F}
    {\sqrt{\left(\|(\bar{H}_s^{(\ell)})^\top \bar{H}_s^{(\ell)}\|_F+\epsilon\right)
    \left(\|(\bar{H}_t^{(\ell)})^\top \bar{H}_t^{(\ell)}\|_F+\epsilon\right)}}.
\label{eq:root-cka}
\end{equation}
Here $\rho_\ell^2$ is the standard linear CKA score in \Cref{eq:cka}. We denote the average CKA mismatch across selected layers as the intermediate alignment term $\mathcal{L}_{\mathrm{CKA}}$.

Rather than using a fixed manually tuned $\lambda$, we balance the intermediate CKA term against the logit loss dynamically. Let $\mathrm{sg}(\cdot)$ denote stop gradient. With the original language model loss skipped during QAD, the optimized objective is:
\begin{equation}
    \mathcal{L}_{\mathrm{total}} =
    \mathcal{L}_{\mathrm{TopK}\text{-}\mathrm{KL}} +
    \mathrm{sg}\!\left(
    \frac{\mathcal{L}_{\mathrm{TopK}\text{-}\mathrm{KL}}}
    {\mathcal{L}_{\mathrm{CKA}}+\epsilon}
    \right)\mathcal{L}_{\mathrm{CKA}}.
\label{eq:loss}
\end{equation}

\subsection{Linear CKA Formulation and Layer Selection}
\label{sec:why-cka}
\label{sec:layer-sel}

We adopt the linear variant of CKA~\cite{kornblith2019similarity} due to its invariance to orthogonal transformations and isotropic scaling, which is critical for \nvfp{} activations that exhibit block wise scale shifts. We compute CKA through feature space covariance matrices rather than explicitly materializing token-token Gram matrices. Given centered activation matrices $\bar{H}_t$ and $\bar{H}_s$, the reported linear CKA score is:
\begin{equation}
    \cka(\bar{H}_t, \bar{H}_s) =
    \frac{\|\bar{H}_s^\top \bar{H}_t\|_F^2}
    {\|\bar{H}_t^\top \bar{H}_t\|_F \, \|\bar{H}_s^\top \bar{H}_s\|_F}.
\label{eq:method-cka}
\end{equation}
Centering ensures translation invariance and reduces scale dominated gradients.

\paragraph{Layer selection strategy.}
Not all intermediate states are equally suitable for CKA alignment. We deliberately select activations that capture stable, high level semantic representations while avoiding structural or distributional artifacts:
\begin{itemize}[leftmargin=*, nosep]
    \item \textit{Target locations:} We attach hooks to decoder block outputs, i.e., post block states after the block has integrated its attention, MLP, or Mamba update. For Nemotron 3 Nano, the CKA configuration aligns all 52 matched decoder block outputs.
    \item \textit{Exclusions:} We avoid prenormalization activations, attention weight matrices, and KV cache states as alignment targets. Prenorm activations are strongly affected by residual stream scale, normalization statistics, and quantization induced outliers; attention weights are less suitable as a unified representation target, especially across quantized and full precision passes and across architectures with non-attention blocks; and KV cache states introduce autoregressive temporal dependencies that can confound batch wise CKA estimation.
    \item \textit{Hybrid architecture handling:} For hybrid architectures such as Nemotron 3 Nano, block output hooks provide a common alignment boundary for both Transformer and Mamba-style blocks. This enables one to one teacher–student alignment without relying on module specific internal states.
\end{itemize}

\subsection{Engineering Optimizations for Low-Bit Training}
\label{sec:engineering}

Computing CKA across full sequences and many layers introduces non trivial memory and compute overhead. Our implementation keeps the added cost practical through four engineering choices:

\begin{enumerate}[leftmargin=*, nosep]
    \item \textbf{Top-$k$ logit distillation:} The output level KL loss is computed only on the teacher's top 8192 vocabulary entries, avoiding full vocabulary activation of the distillation loss while preserving the most informative probability mass.
    \item \textbf{Hook-based activation capture:} ModelOpt registers forward hooks on matched student and teacher decoder blocks. Teacher activations are produced under \texttt{torch.no\_grad()} and detached before loss computation; student activations remain in the autograd graph.
    \item \textbf{Feature-space CKA:} CKA is computed from $\bar{H}^\top\bar{H}$ and $\bar{H}_s^\top\bar{H}_t$ matrices, avoiding explicit construction of $N \times N$ token Gram matrices. The training implementation uses all tokens from the captured microbatch at every optimization step.
    \item \textbf{Dynamic loss balancing:} The intermediate CKA loss is averaged across configured layer pairs and scaled by the stop gradient ratio between logit KL and CKA. This keeps the two objectives on comparable numerical scales without a handtuned global $\lambda$.
\end{enumerate}

These choices preserve the standard QAD training flow while adding geometry aware supervision at decoder block boundaries.

\section{Experiments}
\label{sec:experiments}

\subsection{Experimental Setup}
\label{sec:exp}

\paragraph{Models and quantization.}
We evaluate \ckaqad{} on \nemotron{}~\cite{xin2026qad}, a hybrid Mamba--Transformer MoE model post trained with multi stage reinforcement learning, and \qwen{}~\cite{yang2025qwen3}, a compact Transformer reasoning model. Following the \nvfp{} configuration in the original report, we quantize all GEMM layers in \nemotron{} to \nvfp{} except for the 6 self attention layers and their preceding Mamba2 blocks, which remain in BF16 to stabilize the PTQ baseline. For \qwen{}, we quantize Transformer attention and FFN GEMMs to \nvfp{} and evaluate BF16, \nvfp{}, and \nvfp{}+\ckaqad{} checkpoints on the same benchmark protocols. KV-cache tensors are quantized to FP8 when supported.

\paragraph{Training configuration.}
We use the same QAD pipeline as~\cite{xin2026qad}: the BF16 teacher is frozen, the original language model loss is skipped, and the \nvfp{} student is trained with teacher top-$k$ KL divergence ($k=8192$) on a mixture of cold-start SFT data and RL-generated responses ($\sim$2.5B tokens). For \ckaqad{}, we attach CKA hooks to matched decoder block outputs, compute the all-token CKA alignment term at each optimization step, and rely on dynamic loss balancing to match the CKA term to the logit loss scale. The Nemotron CKA runs use global batch size 64, learning rate $5\mathrm{e}{-}6$, cosine decay to $5\mathrm{e}{-}7$, and maximum sequence length 8192.

\paragraph{Evaluation benchmarks.}
We report primary results on three challenging reasoning and coding benchmarks:
\begin{itemize}[leftmargin=*, nosep]
    \item \textbf{AIME25}: American Invitational Mathematics Examination 2025, measuring mathematical reasoning with chain-of-thought prompting~\cite{aime2025}.
    \item \textbf{GPQA-D}: Graduate-level Google-Proof Q\&A (Diamond subset), testing expert level knowledge~\cite{rein2023gpqa}.
    \item \textbf{LiveCodeBench-v5}: Contamination free code generation and debugging tasks~\cite{jain2024livecode}.
\end{itemize}
For each benchmark, we follow the evaluation protocol in~\cite{xin2026qad}: multiple sampling runs per problem (16 for AIME, 8 for LiveCodeBench/GPQA).

\paragraph{CKA measurement.}
We compute reported linear CKA on centered decoder block activations ($H - \mathbb{E}_{\mathrm{tokens}}[H]$) using the standard score in \Cref{eq:method-cka}. We report both layerwise CKA for analysis and average CKA across the configured block output probes for quantitative comparison. All CKA scores are normalized to $[0,1]$, where 1.0 indicates perfect alignment with the BF16 teacher.

\subsection{Main Results: Accuracy and Representational Alignment}
\label{sec:results}

\Cref{tab:main} summarizes the primary comparison between BF16, PTQ, standard QAD (KL-only), and \ckaqad{}.

\begin{table}[t]
\centering
\caption{Main results on Nemotron 3 Nano. \ckaqad{} recovers both benchmark accuracy and internal representational similarity. AIME25 reports pass@1 averaged over 16 samples per problem. Average CKA is computed over configured decoder block output probes; last layer CKA is measured at the final matched decoder block output.}
\label{tab:main}
\resizebox{\linewidth}{!}{
\begin{tabular}{lccccc}
\toprule
\textbf{Method} & \textbf{AIME25} & \textbf{GPQA-D} & \textbf{LiveCodeBench} & \textbf{Avg. CKA $\uparrow$} & \textbf{Last-layer CKA $\uparrow$} \\
\midrule
BF16 (Teacher) & \textbf{89.1} & \textbf{73.0} & \textbf{72.1} & 1.000 & 1.000 \\
\nvfp{} PTQ & 85.0 & 71.6 & 68.9 & 0.993 & 0.983 \\
QAD (KL-only) & 87.9 & 72.7 & 68.9 & 0.958 & 0.740 \\
\textbf{\ckaqad{}} & \textbf{88.3} & \textbf{72.7} & \textbf{70.3} & \textbf{0.994} & \textbf{0.985} \\
\bottomrule
\end{tabular}
}
\end{table}

\begin{figure}[t]
\centering
\IfFileExists{figures/cka_layerwise.pdf}{
    \includegraphics[width=0.95\linewidth]{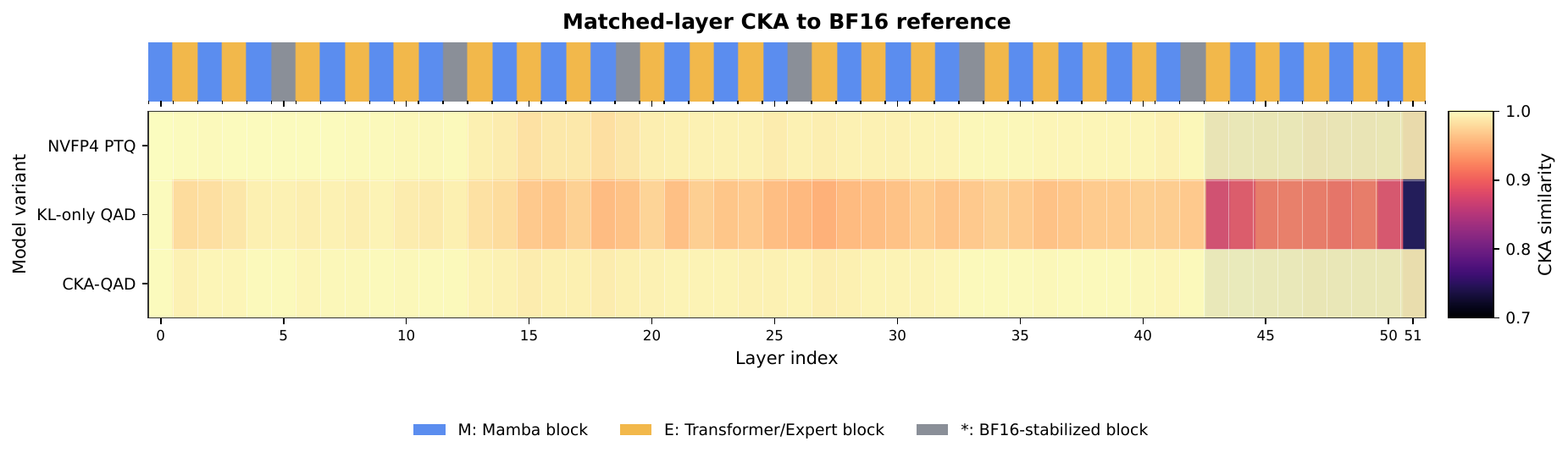}
}{
    \fbox{\parbox{0.95\linewidth}{\centering\textit{Placeholder: figures/cka\_layerwise.pdf}}}
}
\caption{Layer-wise CKA comparison across Nemotron 3 Nano. Standard QAD (KL-only) exacerbates representational drift in mid-depth layers (highlighted), while \ckaqad{} restores alignment across the network. Shaded regions indicate Transformer vs. Mamba blocks.}
\label{fig:cka_layers}
\end{figure}

The results highlight two trends:

\begin{enumerate}[leftmargin=*, nosep]
    \item \textbf{Representational drift in standard QAD.} Despite recovering most benchmark accuracy, QAD (KL-only) exhibits a lower average CKA (0.958) than PTQ (0.993), and its last layer CKA drops sharply to 0.740. This confirms our hypothesis that output level alignment can mask internal feature reorganization. The drift is most pronounced in mid-depth Transformer FFN layers, Mamba state-update blocks, and the final decoder block output (see \cref{fig:cka_layers}).
    \item \textbf{CKA regularization restores internal geometry.} \ckaqad{} raises average CKA from 0.958 under KL-only QAD to 0.994 and last layer CKA from 0.740 to 0.985.
\end{enumerate}

\Cref{fig:cka_layers} visualizes layerwise CKA scores. We observe that:
\begin{itemize}[leftmargin=*, nosep]
    \item PTQ induces moderate CKA degradation uniformly across layers.
    \item QAD (KL-only) further depresses CKA in layers 12--24 (mid-depth Transformer FFN) and layers 35--42 (Mamba state blocks), suggesting these regions are most susceptible to shortcut learning under logit level supervision.
    \item \ckaqad{} restores alignment across all layers, with the largest gains in the aforementioned drift prone regions.
\end{itemize}

\subsection{Compact-Model Validation on \qwen{}}
\label{sec:qwen}

To test whether the downstream behavior extends beyond the hybrid Mamba--Transformer structure of \nemotron{}, we evaluate \qwen{} under BF16, \nvfp{}, and the available \nvfp{}+\ckaqad{} checkpoint. This setting removes Mamba state-update blocks and changes the model scale, serving as a compact Transformer validation of the recovery trends observed on Nemotron. \Cref{tab:qwen} reports the aggregate benchmark results, while \Cref{tab:qwen-lcb} expands the LiveCodeBench-v5 difficulty breakdown.

\begin{table}[t]
\centering
\caption{Average benchmark results and CKA alignment on \qwen{}. AIME25 averages over 16 passes, GPQA-D averages over eight passes, and LiveCodeBench-v5 reports the saved avg@8 accuracy when available. Average CKA is computed over layers 0--35; last-layer CKA uses layer 35.}
\label{tab:qwen}
\resizebox{\linewidth}{!}{
\begin{tabular}{lccccc}
\toprule
\textbf{Method} & \textbf{AIME25} & \textbf{GPQA-D} & \textbf{LiveCodeBench-v5} & \textbf{Avg. CKA $\uparrow$} & \textbf{Last-layer CKA $\uparrow$} \\
\midrule
BF16 (Teacher) & \textbf{79.6} & \textbf{63.2} & \textbf{62.2} & 1.00 & 1.00 \\
\nvfp{} PTQ & 62.9 & 58.8 & 55.4 & 1.00 & 0.99 \\
QAD (KL-only) & 68.5 & 59.5 & 57.9 & 0.98 & 0.86 \\
\textbf{\ckaqad{}} & \textbf{72.3} & \textbf{61.1} & \textbf{59.8} & \textbf{0.99} & \textbf{0.97} \\
\bottomrule
\end{tabular}
}
\end{table}

\begin{table}[t]
\centering
\caption{LiveCodeBench-v5 difficulty breakdown on \qwen{}. BF16 values are averaged over eight passes; \nvfp{} values are the saved single pass averages.}
\label{tab:qwen-lcb}
\begin{tabular*}{\linewidth}{@{\extracolsep{\fill}}lcccc@{}}
\toprule
\textbf{Method} & \textbf{Overall} & \textbf{Easy} & \textbf{Medium} & \textbf{Hard} \\
\midrule
BF16 (Teacher) & \textbf{62.2} & 97.9 & \textbf{74.4} & \textbf{33.9} \\
\nvfp{} PTQ & 55.4 & \textbf{98.3} & 66.0 & 24.4 \\
\bottomrule
\end{tabular*}
\end{table}

The Qwen results show that \nvfp{} quantization substantially reduces average single pass accuracy relative to the BF16 teacher on AIME25 (79.6\% to 62.9\%), GPQA-D (63.2\% to 58.8\%), and LiveCodeBench-v5 (62.2\% to 55.4\%). QAD (KL-only) improves average single pass accuracy to 68.5\% on AIME25, 59.5\% on GPQA-D, and 57.9\% on LiveCodeBench-v5, while reducing average CKA from 1.00 under PTQ to 0.98 and last-layer CKA from 0.99 to 0.86. The completed \qwen{} \ckaqad{} runs reach 72.3\% on AIME25, 61.1\% on GPQA-D, and 59.8\% on LiveCodeBench-v5 while restoring average CKA to 0.99 and last layer CKA to 0.97. The LiveCodeBench-v5 breakdown shows that most of the \nvfp{} degradation is concentrated in medium and hard problems, while easy problem accuracy remains saturated.

\subsection{Training Efficiency and Overhead}
\label{sec:overhead}

We profile \ckaqad{} to quantify engineering overhead under a matched token setup. \Cref{tab:overhead} reports a direct comparison between standard \nvfp{} QAD and \ckaqad{} on \qwen{} using the same H100 training configuration.

\begin{table}[h]
\centering
\caption{Training efficiency of standard \nvfp{} QAD and \ckaqad{} on \qwen{}. Both runs use the same token setup on H100 GPUs.}
\label{tab:overhead}
\resizebox{\linewidth}{!}{
\begin{tabular}{lccc}
\toprule
\textbf{Metric} & \textbf{\nvfp{} QAD} & \textbf{\nvfp{} + \ckaqad{}} & \textbf{Relative Change} \\
\midrule
Wall-clock time per step & 9.822s & 9.871s & +0.5\% \\
Peak VRAM allocated & 73,881 MiB & 79,047 MiB & +7.0\% \\
Throughput (global tokens/sec) & 53,382 & 53,114 & -0.5\% \\
Steps to validation KL $<0.04$ & 4,500 & 4,350 & -3.3\% \\
\bottomrule
\end{tabular}
}
\end{table}

\section{Discussion and Analysis}
\label{sec:disc}

\subsection{The Disconnect Between Output Alignment and Internal Geometry}

Our empirical analysis reveals a counterintuitive phenomenon: while standard QAD successfully minimizes output level KL divergence, it frequently exacerbates internal representational drift compared to PTQ alone. We attribute this disconnect to two intertwined mechanisms.

\paragraph{Shortcut learning via logit level supervision.}
The KL divergence loss $\mathcal{L}_{\mathrm{KL}}$ constrains only the final output distribution $p(y|x)$, leaving the intermediate transformation paths unconstrained. This creates a high-dimensional optimization manifold where the quantized student can satisfy the distillation objective through internal remapping or shortcut learning: reorganizing hidden activations into degenerate or superficial manifolds that happen to project correctly onto the output logits. Such remapping destroys the pairwise similarity structure captured by CKA, as the student no longer needs to preserve the teacher's semantic hierarchy to minimize KL.

\paragraph{Capacity compression under NVFP4 quantization.}
\nvfp{}'s aggressive 4-bit precision and fine-grained block scaling (16-element blocks with E4M3 scales) significantly compress the model's representational capacity. Under this bottleneck, gradient-based optimization naturally favors solutions that minimize loss with minimal representational overhead. Consequently, the student tends to perform early layer feature reorganization to quickly satisfy KL constraints, while deeper layers lose their role in maintaining long range semantic consistency. This effect is particularly pronounced in RL-post-trained models, where the teacher's internal geometry has been heavily specialized through multi-step reasoning rollouts. KL-only QAD fails to preserve this specialized topology, leading to the observed CKA collapse in mid-to-deep Transformer FFN and Mamba state blocks.

\subsection{Why CKA Regularization Enhances Complex Reasoning}

Integrating CKA into the QAD objective directly addresses the geometric degradation described above and improves or preserves downstream accuracy across completed evaluations. We identify two complementary mechanisms:

\paragraph{Preservation of representational topology.}
Linear CKA matches the centered Gram matrices of teacher and student activations, effectively aligning the pairwise similarity structure of the activation space rather than pointwise magnitudes. This acts as a geometric anchor, discouraging collapse into shortcut solutions and encouraging the student to maintain the teacher's manifold structure. By preserving the relative positioning of semantic concepts across layers, CKA regularization helps quantization retain the fine-grained relational knowledge encoded during post training.

\paragraph{Stabilizing multi-step feature propagation.}
Complex reasoning and code generation require coherent feature propagation through dozens of sequential layers. When internal geometry drifts, representational errors can compound, degrading the model's ability to maintain state across long reasoning chains. CKA alignment mitigates this effect by enforcing layerwise consistency, which improves or preserves downstream accuracy across completed evaluations. Our empirical correlation between CKA recovery and benchmark outcomes supports this hypothesis while avoiding the stronger claim that higher CKA must always imply higher task accuracy.

\subsection{Positioning Against Traditional Feature Distillation}

While CKA regularization shares conceptual similarities with feature level distillation (e.g., FitNet~\cite{romero2014fitnets}, SP~\cite{tung2019similarity}), it addresses fundamental limitations that have historically prevented feature distillation from succeeding in low bit settings.

\paragraph{Invariance to non-stationary quantization distributions.}
\nvfp{} activations exhibit highly non-stationary statistics due to per block E4M3 scaling and second level FP32 tensor scaling. CKA is explicitly invariant to orthogonal transformations and isotropic scaling, and its centering operation removes translation bias. This allows CKA to focus on the relational geometry of activations, making it robust to the scale fluctuations inherent in \nvfp{} forward passes.

\paragraph{Optimization stability under severe precision bottlenecks.}
CKA's normalization by Frobenius norms naturally balances the scale of the regularization term, enabling smooth joint optimization of $\mathcal{L}_{\mathrm{KL}}$ and $\mathcal{L}_{\mathrm{CKA}}$ without manual loss weighting schedules. In our setting, this provides a stable way to add representation-level supervision to \nvfp{} QAD while keeping the output level KL objective intact.

\subsection{Limitations and Future Directions}

Despite its effectiveness, \ckaqad{} introduces several practical considerations. First, computing layerwise covariance matrices incurs additional memory and FLOPs, though top-$k$ logit distillation, no gradient teacher forwarding, and featurespace CKA keep overhead below $15\%$ in our runs. Second, CKA alignment is currently applied to static block-output probes; dynamically selecting drift-prone layers based on online CKA monitoring could further improve efficiency. Third, while CKA preserves second-order statistics, it does not explicitly enforce higher-order manifold properties (e.g., curvature or local density), which may become relevant for sub-4-bit formats.

Future work will explore: (1) adaptive CKA balancing based on per layer quantization error or online representation drift; (2) theoretical bounds linking CKA alignment to downstream generalization in quantized networks; and (3) extension of geometry-aware distillation to INT4, MXFP4, and hybrid precision regimes. We also plan to investigate whether CKA regularization can be integrated into the RL training pipeline itself to mitigate representation shifts that later amplify under quantization.

\section{Conclusion}
\label{sec:concl}

In this work, we systematically investigated the internal representational dynamics of \nvfp{}-quantized large language models during quantization-aware distillation (QAD). While standard QAD effectively recovers output level accuracy via KL divergence, we uncovered a critical failure mode: optimizing logit alignment alone frequently exacerbates internal representational drift, as quantified by Centered Kernel Alignment (CKA). This geometric collapse is particularly pronounced in RL-post-trained models, where the student learns superficial mappings that satisfy the distillation objective at the expense of semantic topology. To address this, we proposed \textbf{\ckaqad{}}, a CKA-guided representational alignment method for \nvfp{} QAD and low bit LLM accuracy recovery. Experiments on Nemotron 3 Nano and \qwen{} demonstrate that \ckaqad{} restores internal similarity toward BF16 levels on the primary CKA analysis and improves or preserves downstream accuracy across completed evaluations, all with modest training overhead ($<15\%$).

Our findings support a focused principle for low bit LLM compression: functional alignment at the output layer is necessary but insufficient for faithful capability recovery. Preserving the teacher's internal representational geometry is especially important for tasks requiring deep, multi-step feature propagation such as mathematical reasoning and code synthesis. By explicitly constraining the student's activation space, CKA regularization reduces shortcut learning and helps aggressive quantization retain the fine-grained semantic hierarchies encoded during post training. This work positions CKA-guided representational alignment as a practical extension of output driven QAD for low bit LLM recovery.

Several promising directions remain for future exploration. First, integrating adaptive CKA balancing based on per layer quantization sensitivity or online CKA monitoring could further optimize the accuracy overhead trade off. Second, hardware friendly, low rank, or block diagonal CKA approximations could reduce the overhead of CKA-guided training. Third, extending representational alignment to other low precision formats, including INT4, MXFP4, and sub-4-bit regimes, will rigorously test the generality of CKA-guided low bit recovery. Finally, incorporating CKA-aware constraints directly into reinforcement learning or multi-stage post training pipelines may mitigate representation shifts that later amplify under quantization, offering a unified pathway for robust, high fidelity LLM compression.

We will release our \ckaqad{} implementation, diagnostic tools, and evaluation recipes as lightweight components compatible with Megatron-LM, NeMo, and HuggingFace workflows, with the goal of accelerating CKA-guided representational alignment for production low bit LLM deployment.

\bibliographystyle{plain}

\begin{thebibliography}{99}

\bibitem{xin2026qad}
Meng Xin, Sweta Priyadarshi, Jingyu Xin, et al.
\newblock Quantization-aware distillation for NVFP4 inference accuracy recovery.
\newblock arXiv:2601.20088, 2026.

\bibitem{alvarez2025nvfp4}
Eduardo Alvarez, Omri Almog, Eric Chung, Simon Layton, Dusan Stosic, Ronny Krashinsky, and Kyle Aubrey.
\newblock Introducing NVFP4 for efficient and accurate low-precision inference.
\newblock NVIDIA Technical Blog, 2025.

\bibitem{kornblith2019similarity}
Simon Kornblith, Mohammad Norouzi, Honglak Lee, and Geoffrey Hinton.
\newblock Similarity of neural network representations revisited.
\newblock In ICML, 2019.

\bibitem{elhage2022superposition}
Nelson Elhage, Tristan Hume, Catherine Olsson, et al.
\newblock Toy models of superposition.
\newblock Transformer Circuits Thread, 2022.

\bibitem{dao2024mamba2}
Tri Dao and Albert Gu.
\newblock Transformers are SSMs: Generalized models and efficient algorithms through structured state space duality.
\newblock In ICML, 2024.

\bibitem{yang2025qwen3}
An Yang, Anfeng Li, Baosong Yang, Beichen Zhang, Binyuan Hui, Bo Zheng, Bowen Yu, Chang Gao, Chengen Huang, Chenxu Lv, Chujie Zheng, Dayiheng Liu, Fan Zhou, Fei Huang, Feng Hu, Hao Ge, Haoran Wei, Huan Lin, Jialong Tang, Jian Yang, Jianhong Tu, et al.
\newblock Qwen3 technical report.
\newblock arXiv:2505.09388, 2025.

\bibitem{nagel2021white}
Markus Nagel, Marios Fournarakis, Rana Ali Amjad, Yelysei Bondarenko, Mart van Baalen, and Tijmen Blankevoort.
\newblock A white paper on neural network quantization.
\newblock arXiv:2106.08295, 2021.

\bibitem{hinton2015distilling}
Geoffrey Hinton, Oriol Vinyals, and Jeff Dean.
\newblock Distilling the knowledge in a neural network.
\newblock arXiv:1503.02531, 2015.

\bibitem{romero2014fitnets}
Adriana Romero, Nicolas Ballas, Samira Ebrahimi Kahou, Antoine Chassang, Carlo Gatta, and Yoshua Bengio.
\newblock FitNets: Hints for thin deep nets.
\newblock arXiv:1412.6550, 2014.

\bibitem{zagoruyko2017attention}
Sergey Zagoruyko and Nikos Komodakis.
\newblock Paying more attention to attention: Improving the performance of convolutional neural networks via attention transfer.
\newblock In ICLR, 2017.

\bibitem{tung2019similarity}
Frederick Tung and Greg Mori.
\newblock Similarity-preserving knowledge distillation.
\newblock In ICCV, 2019.

\bibitem{jung2023feature}
Hee-Jun Jung, Doyeon Kim, Seung-Hoon Na, and Kangil Kim.
\newblock Feature structure distillation with Centered Kernel Alignment in BERT transferring.
\newblock Expert Systems with Applications, 234:120980, 2023.

\bibitem{zhou2024rethinking}
Zikai Zhou, Yunhang Shen, Shitong Shao, Linrui Gong, and Shaohui Lin.
\newblock Rethinking Centered Kernel Alignment in Knowledge Distillation.
\newblock In IJCAI, 2024.

\bibitem{aime2025}
Mathematical Association of America.
\newblock American Invitational Mathematics Examination 2025.
\newblock 2025.

\bibitem{rein2023gpqa}
David Rein, Betty Li Hou, Asa Cooper Stickland, Jackson Petty, Richard Yuanzhe Pang, Julien Dirani, Julian Michael, and Samuel R. Bowman.
\newblock GPQA: A graduate-level Google-proof Q\&A benchmark.
\newblock arXiv:2311.12022, 2023.

\bibitem{jain2024livecode}
Naman Jain, King Han, Alex Gu, Wen-Ding Li, Fanjia Yan, Tianjun Zhang, Sida Wang, Armando Solar-Lezama, Koushik Sen, and Ion Stoica.
\newblock LiveCodeBench: Holistic and contamination free evaluation of large language models for code.
\newblock arXiv:2403.07974, 2024.

\end{thebibliography}

\end{document}